\pdfoutput=1

\documentclass[11pt]{article}

\usepackage[]{acl}
\usepackage{authblk}

\usepackage{times}
\usepackage{latexsym}
\usepackage{tcolorbox}

\usepackage{amsmath}
\usepackage{subcaption}

\usepackage{booktabs}
\usepackage{amssymb}
\usepackage{graphicx}
\usepackage[frozencache,cachedir=.]{minted}

\usepackage{multirow}

\newlength{\myMheight}
\newcommand{\hf}{\includegraphics[height=\myMheight]{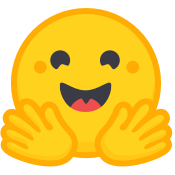}}

\newcommand{\tightparagraph}[1]{%
  \paragraph{#1}%
  \vspace{-0.5\baselineskip}%
}
\usepackage{tcolorbox}
\tcbuselibrary{skins}
\newcommand{\improvedParbox}[1]{%
  \begin{tcolorbox}[
    enhanced,
    colback=white,
    colframe=gray!50!black,
    boxrule=0.5pt,
    arc=1mm,
    left=3mm,
    right=3mm,
    top=2mm,
    bottom=2mm,
    boxsep=0pt,
    width=\columnwidth
  ]
    #1
  \end{tcolorbox}
}

\usepackage[T1]{fontenc}
\usepackage[utf8]{inputenc}
\usepackage{microtype}

\title{
Attention Overflow: Language Model Input Blur\\
during Long-Context Missing Items Recommendation 
}

\author[]{Damien Sileo}
\affil[]{Univ. Lille, Inria, CNRS, Centrale Lille, UMR 9189 - CRIStAL, F-59000 Lille, France}
\affil[]{\url{damien.sileo@inria.fr}}

\begin{document}
\maketitle
\begin{abstract}
Large language models (LLMs) can suggest missing elements from items listed in a prompt, which can be used for list completion or recommendations based on users' history. However, their performance degrades when presented with too many items, as they start to suggest items already included in the input list. This occurs at around 100 items for mid-2024 flagship LLMs. We evaluate this phenomenon on both synthetic problems (e.g., finding missing numbers in a given range of shuffled integers) and realistic movie recommendation scenarios. We refer to this issue as \textit{attention overflow}, as preventing repetition requires attending to all items simultaneously. Although iterative loops can mitigate this problem, their costs increase with the repetition rate, affecting the language models' ability to derive novelty from lengthy inputs.

\end{abstract}

\section{Introduction}

Large language models (LLMs) boast ever-growing context windows, enabling new potential applications. However, the theoretical context length is not a sufficient indication of a model's real performance with a given input size \cite{liu2024lost}. Multiple benchmarks have been proposed to stress-test the actual capabilities of language models to reason over long contexts. Most of these tasks are either pure retrieval or involve a form of reasoning, requiring the identification of a few relevant portions from a large context.

We question the effective context window of language models from an opposite angle: asking them to provide the only relevant elements that are \textit{not} in a large input. We formulate this as a missing item prediction task. Missing item prediction has multiple applications, notably in conversational recommendation, where users can provide a list of movies they have already watched and ask for new suggestions. This task involves a form of inductive reasoning, in contrast to the deductive reasoning typically explored in long context stress tests. More importantly, it requires comparing a representation to the whole input, and we notice that this is difficult for current LLMs, which leads to the prediction of items already in the input (repetition).

Missing item prediction is relevant when models are asked to generate long lists, as we have observed repetitions in this scenario\footnote{For example, asking Claude Sonnet 3.5 200 movies released in 2022 leads to numerous repetitions: \href{https://claude.site/artifacts/67f091d2-4ab5-4b88-9fce-b4114ade666e}{[artifact]}}, but we focus on the movie recommendation use case, where users provide the movies they have watched, and we also create synthetic examples, notably number ranges with a missing element. We quantify the repetition phenomenon with existing off-the-shelf language models and investigate whether fine-tuning can easily address this problem. The generated datasets are publicly available\footnote{
\href{https://huggingface.co/datasets/sileod/missing-item-prediction}{[data:HF-datasets \hf]}\label{footnote:urls}}.

\section{Related work}

\begin{figure*}[htb]
    \centering
    \begin{subfigure}{\textwidth}
        \centering
        \includegraphics[trim= 0 0 140 0,clip,width=\textwidth]{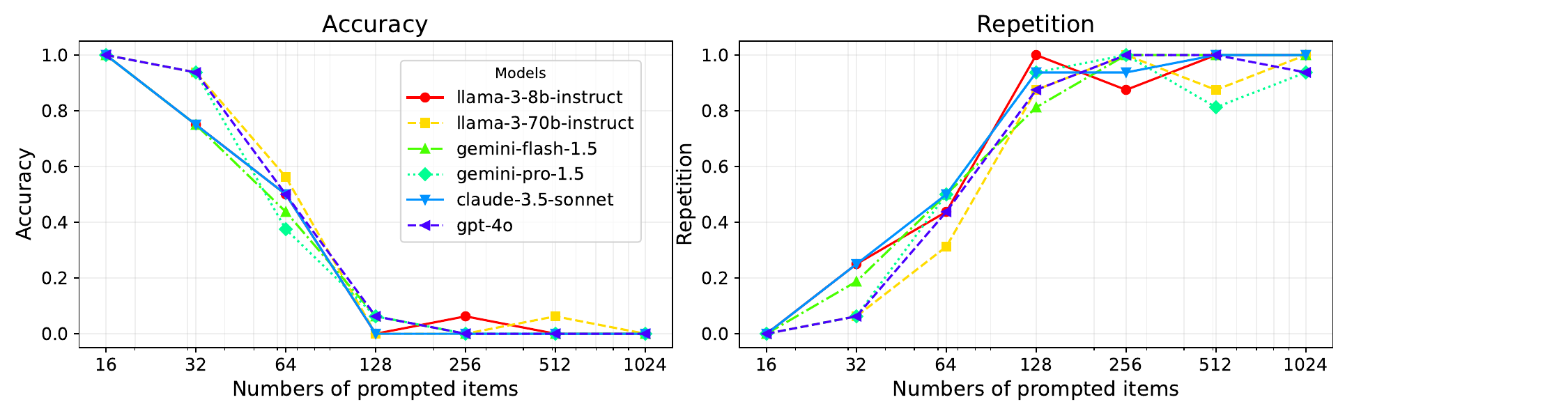}
        \caption{Zero-shot missing number prediction}
    \end{subfigure}
    \begin{subfigure}{\textwidth}
        \centering
        \includegraphics[trim= 0 0 140 0,clip,width=\textwidth]{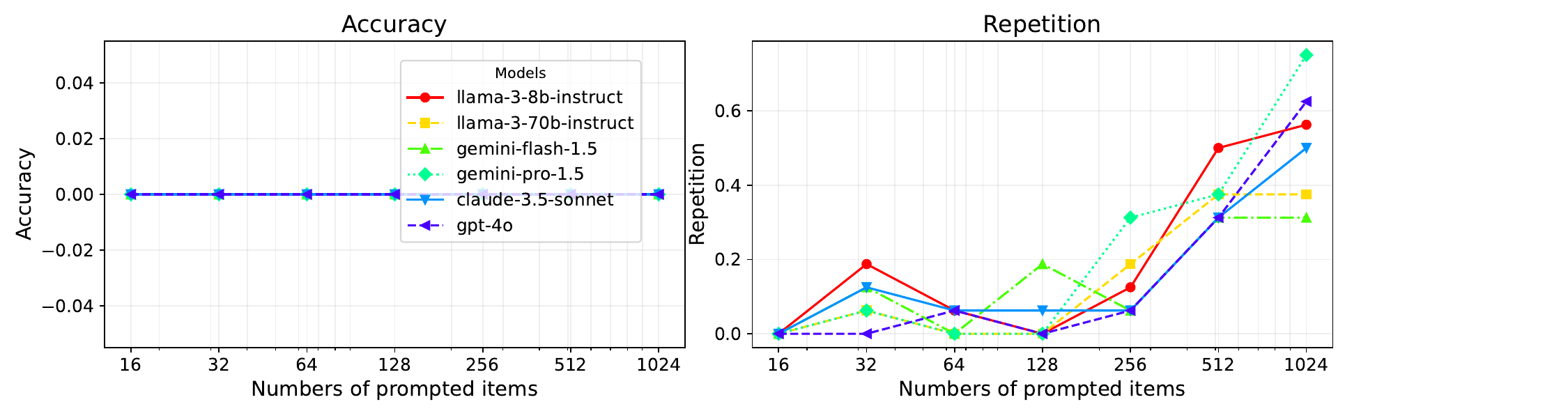}
        \caption{Zero-shot missing movie prediction}
    \end{subfigure}
    \caption{Zero-shot test accuracy and repetition rate with increasing itemset sizes.}
    \label{fig:zs}
\end{figure*}

\paragraph{Repetitions in language modeling} We study a form of repetitions, a well-identified problem in language models \cite{keskar2019ctrl}, which can sometimes lead to text degeneration, where models repeat the same token indefinitely \cite{fu2021theoretical}. Repetition penalties were proposed to alleviate this issue \cite{keskar2019ctrl}, but they operate at the token level and cannot scale to large contexts where all tokens are already represented. Repetitions also exist in more subtle ways, as \citet{chiang-lee-2024-reasoning} showed that chain-of-thought reasoning contains redundant content.

\paragraph{LLM context length stress tests}  Our work is also related to context window stress testing and language modeling-based recommendation.
Previous work has studied the ability of attention mechanisms to identify what is present in long contexts, but not what is missing. The Long-Range Arena \cite{tay2021long} provides the first systematic analysis of the long-range processing capabilities of text encoders, focusing mainly on algorithmic reasoning and retrieval tasks. BABILong  \cite{kuratov2024babilong} uses bAbi reasoning tasks \cite{babi} and interleaves relevant text with irrelevant input. FlenQA \cite{levy2024same} applies a similar process to the RuleTaker \cite{ruletaker} deductive logical reasoning task.

\paragraph{Recommendation with LLMs}
Our study is also related to LLM usage for collaborative filtering \cite{sileo-lmrec-2022}, where users enumerate a list of items to communicate their tastes. LLMs can also be used in content-based recommendations, where users explicitly mention what they are looking for \cite{wu2023survey}. Here, we do not address the fine-grained relevance of the recommendations (providing an item that users do not already know). Repetition is also related to the novelty metric in recommender systems evaluation \cite{vargas2011rank}.

\section{Missing item prediction}

\begin{figure*}[htb]
    \centering
    \begin{subfigure}{\textwidth}
        \centering
        \includegraphics[trim= 0 0 140 0,clip,width=\textwidth]{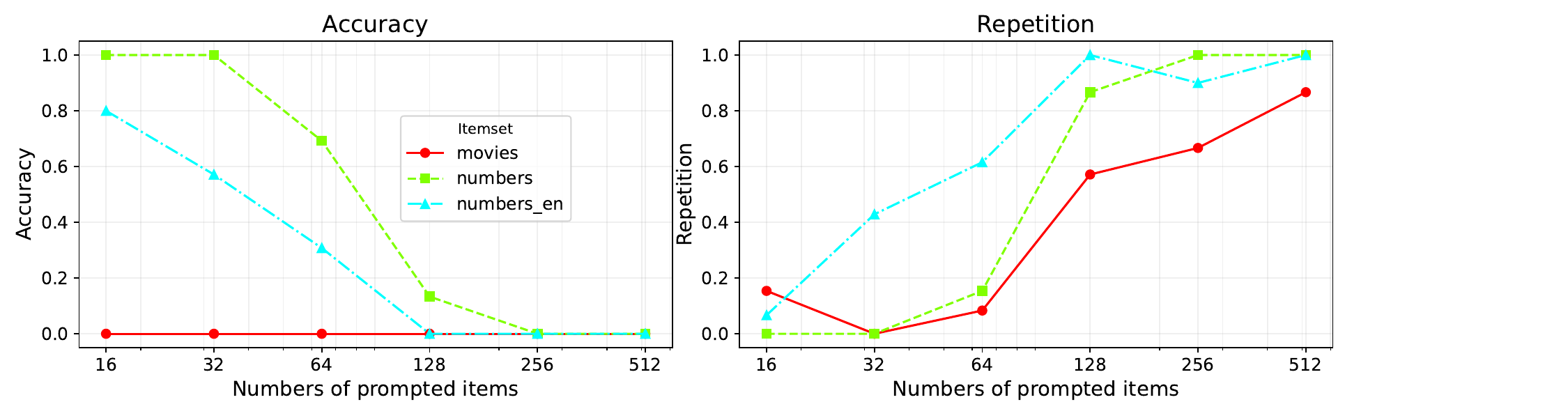}
        \caption{Llama-3 zero-shot missing number prediction on multiple domains}
    \end{subfigure}
    \begin{subfigure}{\textwidth}
        \centering
        \includegraphics[trim= 0 0 140 0,clip,width=\textwidth]{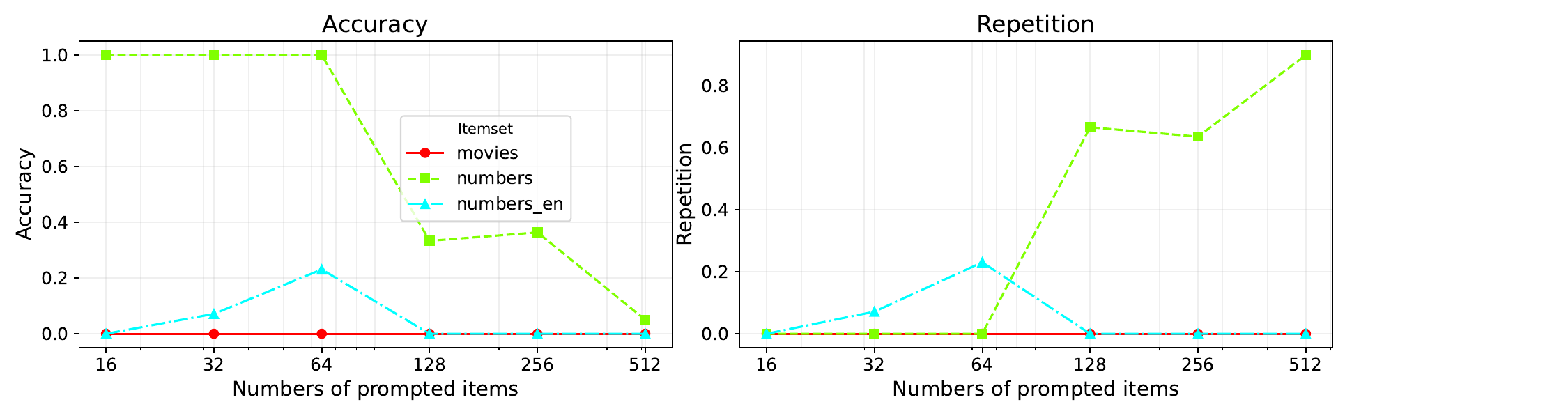}
        \caption{Llama-3 fine-tuned on missing number prediction}
    \end{subfigure}
    \caption{Llama-3-8B-Instruct Accuracy on on various itemsets with increasing itemset sizes, without any fine-tuning (a) and after fine-tuning on the numebers itemset.}
    \label{fig:ft}
\end{figure*}

We formalize the task of missing item prediction as follows: Given a set $X$ (=randomly ordered) of $N$ elements, guess the element $y$ that is missing in $X$.
This is technically an induction task that can be under-determined but we can construct relatively easy $X,y$ pairs with easily identifiable itemsets $\mathcal{S}$ (numbers from $0$ to $1024$, letters, chemical elements...) and randomly removing one element $y$ from $\mathcal{S}$ to get $X$.
We can use two evaluation metrics:
\tightparagraph{Accuracy} the rate at which a language model returns the expected missing element.
\tightparagraph{Repetition rate} the rate at which a language model returns an element that is already in $X$.

Repetitions are always mistakes. For easily identifiable sets, ideal behavior is perfect accuracy and no repetition. But even in cases where the structure of $\mathcal{S}$ is under-determined, language models performing missing item prediction should not repeat elements from $X$.

To construct an example of the missing item prediction task, we select an itemset $\mathcal{S}$, select a random element $y$, and present a scrambled version of $X=\mathcal{S}\setminus \{y\}$ in a prompt explicitly asking the model to guess a missing element. We provide the following itemsets:
\tightparagraph{Movies} We select a user from the MovieLens 1M dataset who watched more than 2048 movies.
\tightparagraph{Numbers} Numbers in numerical form (1...1024). We exclude set extrema from the choice of $y$ for numerical itemsets. 
\tightparagraph{Numbers-english} We use the same numbers but converted in English using the num2word library \footnote{\url{https://github.com/savoirfairelinux/num2words}}.

An example with the Numbers itemset of size 8 is \textsc{Question}: \textit{Find the missing element in  5, 7, 1, 3, 6, 8, 4.} \textsc{Answer:} 2.

\section{Experiments}

We use the same prompt template for all models:
\improvedParbox{
Guess the missing item from this list:
\{X\}. 
Directly answer with only one item. Item format should match the list format. Provide no explanation. Answer format: "\{item\}."
}%

To construct this prompt template, we iterated on Llama-3-8B-Instruct with the numbers itemset validation data until we obtained a satisfactory output format.

We then normalize the outputs with punctuation removal and lowercase to compute repetition rate and accuracy, and perform exact matches to compute accuracy and repetition rate.

We use powers of 2 from 16 starting from 16 as itemsize. This ensures that there are enough items to guess the itemset structure. We generate 200 train examples and 100/100 validation test examples per itemset size and itemset type.

\subsection{Zero-shot evaluation}
We evaluate off-the-shelf instruction-tuned language models API through OpenRouter.
We evaluate Llama3-Instruct 8B and 70B, Gemini 1.5 Flash and Pro, GPT-4o, and Claude 3.5 Sonnet on July 10th 2024, with the default hyperparameters.


Figure \ref{fig:zs} shows the evolution of Accuracy and Repetition metrics with different itemsets sizes for numeric numbers and movies missing item prediction tasks. Most language models solve the missing number prediction task with relatively high accuracy with less than 128 items. Increasing model size seems to improve accuracy, as Gemini Pro and Llama-3-70B outperform their smaller counterpart. However, the repetition rates shoot up and the accuracy decreases in all models after 256 items.

We cannot interpret the low accuracy of the movie item prediction tasks as a failure because the models can predict relevant movies that are not $y$. However, we can interpret the growing repetition rate as a failure, which can frustrate users who could expect better recommendations as they provide more examples, which limits the accuracy of conversational recommender systems that not filtering their output to prevent repetitions.

\subsection{Fine-tuning}
We now investigate whether fine-tuning can easily address this issue. We fine-tune Llama-3 Instruct 8B using Unsloth default configuration
\footnote{\url{https://colab.research.google.com/drive/135ced7oHytdxu3N2DNe1Z0kqjyYIkDXp?usp=sharing}} (4bit quantization, Lora \cite{dettmers2024qlora} with dimension 16, 1 epoch with a learning rate of 2e-4). We fine-tune on 500 numeric items of a size below 256 and evaluate on the test set in-domain and out-domain.

Figure \ref{fig:ft} shows that fine-tuning improves missing item prediction on in-domain data, but do not generalize to larger itemsets nor to different domains, which might indicate a fundamental limit of current attention architectures that may not be solved with data only.

\subsection{Contrastive evaluation}
\begin{figure}[H]
    \includegraphics[width=0.5\textwidth]{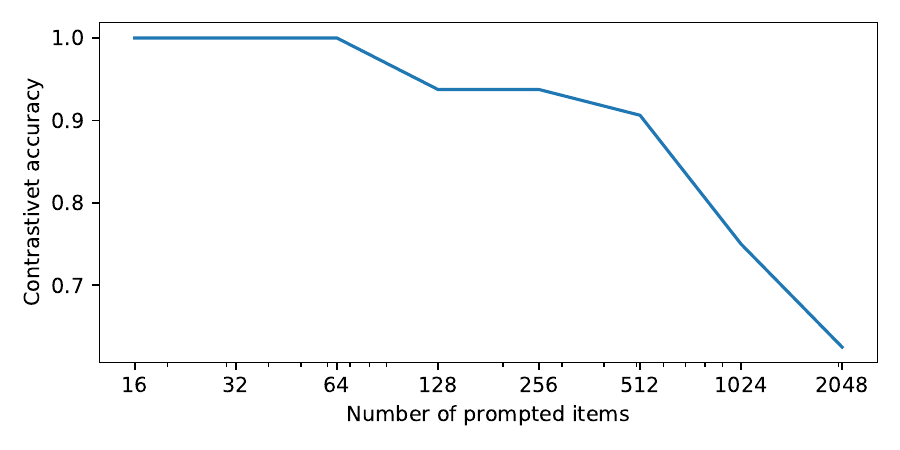}
    \caption{Zero-shot contrastive accuracy with Lllama-3-8B-Instruct on the Numbers itemset.}
    \label{fig:contrastive}
\end{figure}

We also evaluated the ability of LLama-3-8B-Instruct to tell whether an element is present or not in the list by randomly sampling either the missing element or a random element from a prompt. 
\improvedParbox{%
  \{X\}.
  Is "\{i\}" in the previous list?  Provide no explanation, directly answer with only "Yes." or "No."
}
Figure \ref{fig:contrastive} shows the evolution of accuracy with growing itemset sizes. Llama-3-8B-Instruct maintains $75\%$ accuracy below 1024 items\footnote{All examples fit in the 8K context window of Llama 3.}. This shows that once the item is explicitly present in the query, the model is much better at identifying it. These results are lower than the Needle in a Haystack evaluation scores of Llama-3 \cite{zhang2024extending}, which is due to the high similarity between items. This suggests that context-length stress testing is harder when all many prompt elements are similar to each other, and which would make BABILong  \cite{kuratov2024babilong} problem lengthening too easy get around. 

\section{Analysis}
To solve missing item prediction, a transformer language model needs to construct a latent representation of the missing item when predicting the next token. Finding a close representation is relatively simple in the tasks we propose, as language models consistently output items that belong to the item set. However, they also need to compare the latent representation with the latent representations of the prompted items. At each layer, the transformer layer can refine the representation to shift it away from prompted items, but the models lack the depth to do it for many items. 


\section{Conclusion}

We introduce a new missing item prediction dataset and we show that repetitions occur during movie recommendation tasks, which is a real-world problem, alongside list completion. This also has implication on the current language models to check exhaustivity in texts. Our simple examples show that we must be careful when asking language models to produce new content from contextual information, as language models can repeat context elements without noticing it. This finding provides further evidence for the need for caution when interpreting context length \cite{liu2024lost}. We attribute this phenomenon to an overflow of attention, speculating that the model needs to evaluate candidates and compare them to all input items at once. It would be worthwhile to actually analyze the attention heads during this task, even though multi-head attention is hard to interpret \cite{bibal2022attention}. Our dataset is publicly available with itemset sizes up to 8192 for future work.

\bibliography{anthology,custom}

\end{document}